\pdfoutput=1
\documentclass[11pt]{article}

\usepackage[final]{acl}

\usepackage{times}
\usepackage{latexsym}

\usepackage[T1]{fontenc}
\usepackage[utf8]{inputenc}

\usepackage{microtype}

\usepackage{inconsolata}

\usepackage{graphicx}

\title{Towards Fairness Assessment of Dutch Hate Speech Detection}

\author{
    \textbf{Julie Bauer}\textsuperscript{1},
    \textbf{Rishabh Kaushal}\textsuperscript{1,2},
    \textbf{Thales Bertaglia}\textsuperscript{3},
    \textbf{Adriana Iamnitchi}\textsuperscript{1}
    \\
    \\
    \textsuperscript{1}Maastricht University
    \textsuperscript{2}Indira Gandhi Delhi Technical University for Women
    \textsuperscript{3}Utrecht University
    \\
 \small{
   \textbf{Correspondence:} \href{mailto:jtsg.bauer@student.maastrichtuniversity.nl}{jtsg.bauer@student.maastrichtuniversity.nl}
 }
}

\begin{document}
\maketitle
\begin{abstract}
Numerous studies have proposed computational methods to detect hate speech online, yet most focus on the English language and emphasize model development. In this study, we evaluate the counterfactual fairness of hate speech detection models in the Dutch language, specifically examining the performance and fairness of transformer-based models.
We make the following key contributions. First, we curate a list of Dutch Social Group Terms that reflect social context. Second, we generate counterfactual data for Dutch hate speech using LLMs and established strategies like Manual Group Substitution (MGS) and Sentence Log-Likelihood (SLL). Through qualitative evaluation, we highlight the challenges of generating realistic counterfactuals, particularly with Dutch grammar and contextual coherence. Third, we fine-tune baseline transformer-based models with counterfactual data and evaluate their performance in detecting hate speech. Fourth, we assess the fairness of these models using Counterfactual Token Fairness (CTF) and group fairness metrics, including equality of odds and demographic parity. Our analysis shows that models perform better in terms of hate speech detection, average counterfactual fairness and group fairness. This work addresses a significant gap in the literature on counterfactual fairness for hate speech detection in Dutch and provides practical insights and recommendations for improving both model performance and fairness.
\end{abstract}

%%%%%%%%%%%%%%%%%%%%%%%%%%%%%%%%%%%%%%%%%%%%%%
\section{Introduction}
%%%%%%%%%%%%%%%%%%%%%%%%%%%%%%%%%%%%%%%%%%%%%%
\begin{figure*}
    \centering
    \includegraphics[width=0.7\textwidth]{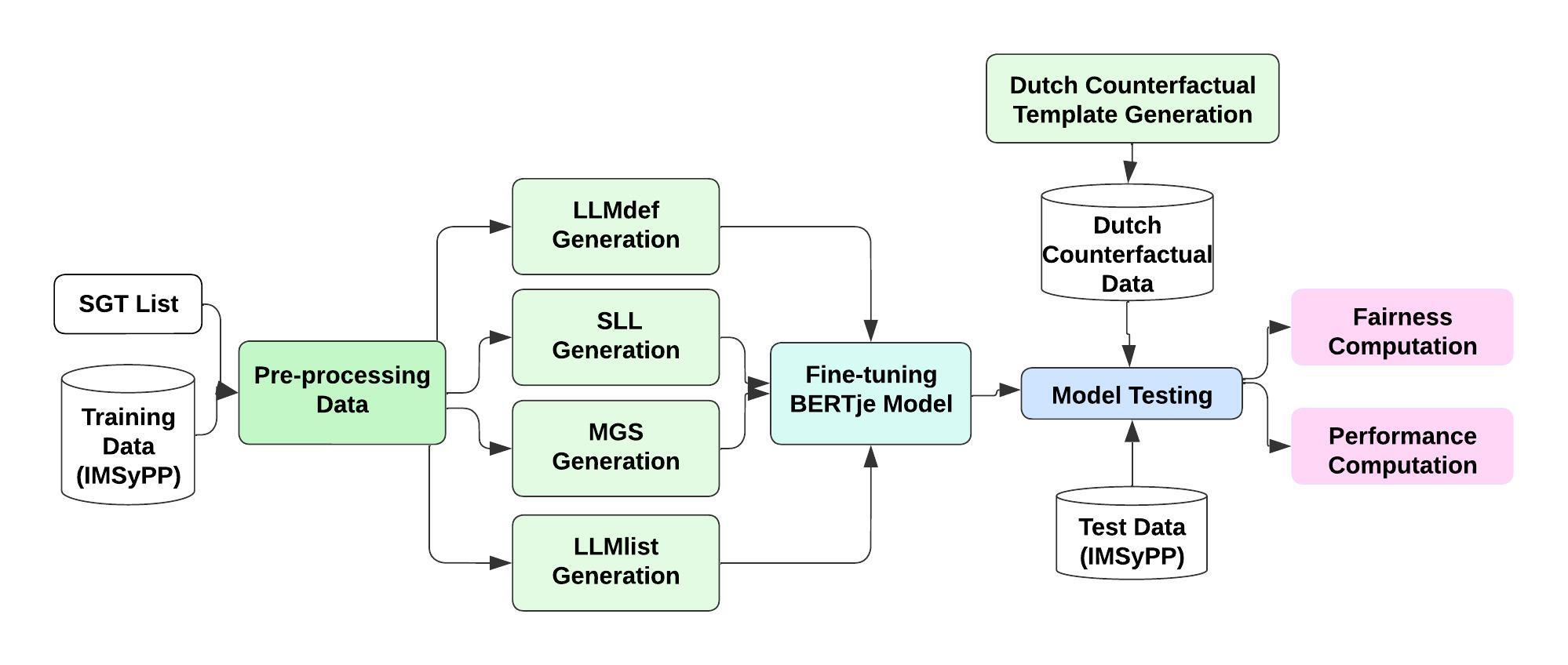} 
    \caption{Proposed methodology outlining key steps. SGT list is curated. Training data forms input to the four counterfactual data generation methods, namely, LLMlist, LLMdef, SLL, and MGS. BERTje model is fine-tuned with counterfactual data. The model is tested on test data and Dutch counterfactual data generated using templates. Finally, performance evaluation and fairness computations are performed.}
    \label{fig:pipeline}
\end{figure*}
While the ease of expressing oneself on social media platforms has led to creative and meaningful interactions, it has also amplified the spread of hate speech -- particularly content targeting specific groups based on ethnicity, gender, sexual orientation, and similar characteristics. 
To address this issue, researchers have developed numerous computational methods for detecting hate speech~\cite{alkomah2022literature,yin2021towards,macavaney2019hate}.  
However, the majority of these efforts focus on the English language. 
Regardless of language, a critical question remains: \textit{Are these detection models fair?} In other words, do they detect hate speech targeted at all social groups with similar accuracy? If not, the models are unfair~\cite{mehrabi2021survey,pessach2022review}. Unfair models risk perpetuating biases, which can exacerbate existing issues and erode users' trust in social media platforms.
This work evaluates the fairness of hate speech detection models for the Dutch language. Consider the following two sentences: S1:\textit{``All Moroccans are troublemakers.''} and S2:\textit{``All Dutch are troublemakers''}. A fair hate speech detection model should classify both sentences as equally hateful since they share the same structure and level of negativity. However, if the model predicts S1 to be 98\% likely to be hateful but S2 only 10\%, this disparity indicates unfairness. This kind of bias often arises when sensitive identity attributes, such as nationality, ethnicity, sexuality, or religion, disproportionately influence model predictions~\cite{garg2019counterfactual}. This example highlights the importance of \textit{counterfactual fairness}: the principle that a model’s decision should remain consistent if sensitive attributes in the input data are changed. For instance, S2 is the counterfactual version of S1 (and vice versa). If a model evaluates these sentences differently, it fails to meet the standard of counterfactual fairness~\cite{kusner2017counterfactual}. This challenge is particularly critical in hate speech detection, where sensitive attributes often appear in potentially biased contexts.
More formally, a model $M$ is considered counterfactually fair if it produces the same predictions for all possible values of a sensitive attribute $A$. Mathematically, this can be expressed as: $ Pr(\hat{Y} \mid A = a_1) = \Pr(\hat{Y} \mid A = a_2)$, where $A \in \{a_1, a_2\}$ represents different values of the sensitive attribute.

Despite progress in counterfactual fairness research for the English language, not much work has been done in European languages such as Dutch. Several studies on Dutch hate speech detection have been conducted using pre-trained language models and mainly focused on creating datasets and models for hate speech detection~\cite{caselli2021dalc,caselli2023benchmarking,lemmens2021improving,markov2022role,markov2022ensemble,ruitenbeek2022zo,theodoridis2022all}. This work aims to bridge the gap in the existing literature by exploring the counterfactual fairness of pre-trained models in detecting hate speech in Dutch on social media platforms. We aim to answer the following research questions: \emph{What are the methods to generate counterfactual data for the Dutch language? Are hate speech detection models in the Dutch language counterfactually fair? What is the impact of generated counterfactual data on the performance and fairness of the hate detection model in the Dutch language?}

Figure~\ref{fig:pipeline} presents the key steps in our methodology.
We begin by curating a list of Dutch Social Group Terms and generating counterfactual sentences. To do this, we use Large Language Models (LLMs) and other techniques like Manual Group Substitution (MGS) and Sentence Log-Likelihood (SLL) originally proposed for English. Through qualitative evaluation, we identify challenges in generating realistic counterfactual sentences that conform to the rules of Dutch grammar. Next, we fine-tune transformer-based hate speech detection models using the generated counterfactual data. We then evaluate the performance of these models in detecting hate speech in Dutch. 
Finally, we assess counterfactual fairness using Counterfactual Token Fairness (CTF).  
We also compute group fairness metrics, namely, equality of odds and demographic parity.

%%%%%%%%%%%%%%%%%%%%%%%%%%%%%%%%%%%%%%%%%%%%%%
\section{Related Work}
%%%%%%%%%%%%%%%%%%%%%%%%%%%%%%%%%%%%%%%%%%%%%%

Hate speech detection has been extensively studied~\cite{macavaney2019hate,bertaglia2021abusive,mullah2021advances, yin2021towards, alkomah2022literature, subramanian2023survey, rawat2024hate, gandhi2024hate}.
Defining hate speech is inherently challenging, as it is a complex phenomenon influenced by interpretation~\cite{hietanen2023towards}.
Fortuna et al.~\cite{fortuna2018survey} proposed a comprehensive definition of hate speech:
\textit{``Hate speech is language that attacks or diminishes, incites violence or hate against groups based on specific characteristics such as physical appearance, religion, descent, national or ethnic origin, sexual orientation, gender identity, or others. It can manifest in various linguistic styles, including subtle forms or even through humour.''}
This definition captures the diverse ways in which hateful language can be expressed.

Although most of the research in this domain has focused primarily on the English language, some studies have explored hate speech detection in multiple languages~\cite{corazza2020multilingual}. Notable examples include investigations into Italian~\cite{del2017hate}, Danish~\cite{sigurbergsson2020offensive}, and Spanish~\cite{plaza2021comparing}, among others.
We focus on hate speech detection in the Dutch language. 
Among earlier works, Tulkens et al.~\cite{tulkens2016automated,tulkens2016dictionary} performed a dictionary-based approach for the detection of racist discourse in Dutch using automated means. Markov et al.~\cite{markov2021exploring} explored features based on emotions and style for cross-domain hate speech detection in multiple languages including Dutch. 
Caselli et al.~\cite{caselli2021dalc} introduced a new dataset, Dutch Abusive Language Corpus (DALC v1.0), which comprises manually annotated tweets for abusive language.
Ruitenbeek et al.~\cite{ruitenbeek2022zo} curated a corpus containing more than 11k posts on Twitter in Dutch which are abusive and offensive. 
Hilte et al.~\cite{hilte2023haters} investigated the demographics of authors who spread hate speech in Dutch and found that older men indulge in more hate speech.
Vries~\cite{vries2024analysing} used a BERT-CNN based model for detecting the targets against whom hate is triggered on the X platform in Dutch.
However, none of these works address the important issue of fairness of Dutch hate speech detection models.

Fairness is becoming increasingly important in the context of hate speech classification. A model is considered fair when it (1)~does not use sensitive attributes in making decisions and (2)~treats individuals with the same sensitive attributes similarly~\cite{mehrabi2021survey}. 
Approaches to increase fairness are predominantly based on sensitive attributes that point to (un)privileged groups, which are disproportionately likely to be positively classified as hateful by a model~\cite{caton2024fairness}. 
Although different notions of fairness exist~\cite{verma2018fairness}, 
we focus on causality-based notions because evaluating causal relationships provides a more comprehensive evaluation of model fairness that can uncover model bias.
%\subsection{Counterfactual Fairness}
More specifically, we apply counterfactual fairness
%, a relatively new concept within the field of machine learning fairness, originally introduced by Kusner et al.
~\cite{kusner2017counterfactual,garg2019counterfactual} which is defined as \textit{``the intuition that a decision is fair towards an individual if it is the same in (a)~the actual world and (b)~a counterfactual world where the individual belonged to a different demographic group''}. 
In the context of fairness of hate speech models~\cite{davani2021improving}, this means that changing the sensitive attribute in a sentence should not impact the outcome of classification. 
The sentences ``some people are gay'' and ``some people are straight'' should, therefore, receive a similar toxicity prediction by fair models. 
While most previous work focuses on English, we focus on the Dutch language and evaluate hate speech detection models from a fairness perspective.

%%%%%%%%%%%%%%%%%%%%%%%%%%%%%%%%%%%%%%%%%%%%%%
\section{Counterfactual Data Generation}
\label{sec:counterf-data}
%%%%%%%%%%%%%%%%%%%%%%%%%%%%%%%%%%%%%%%%%%%%%%

Evaluating the counterfactual fairness of a model requires access to counterfactual sentences. Counterfactual generation is a data augmentation strategy that creates such sentences by modifying sensitive identity terms while preserving the original meaning. This additional data is then used to fine-tune the model with the goal of improving both performance and fairness. We apply three methods for generating counterfactuals: Large Language Model (LLM), Sentence Log-Likelihood (SLL), and Manual Group Substitution (MGS).

\subsection{Dataset Augmentation with Social Groups}
%Hate speech datasets in languages other than English are often limited. 
%We carefully select an appropriate dataset based on accessibility, presence of target groups, and trustworthiness.
We selected the IMSyPP Dutch hate speech dataset\footnote{http://imsypp.ijs.si/}, which is publicly available and was curated as part of a project funded by the European Commission to tackle online hate speech through prevention, awareness, and regulation.
The dataset contains comments from several Dutch social media platforms and forums, such as Twitter or Dumpert, posted from January 2018 to October 2020. 
It consists of a training set with 25,720 posts and an evaluation set with 2,858 posts. %, for a total of 28,578 comments.
The records in the dataset include the website URL that the post originates from, the text of the hate speech post, the hate speech target and the type of hate speech (appropriate, inappropriate, offensive, or violent). 
Of these labels, the `offensive' and `violent' classes are typically regarded as hate speech. 
The target categories are one of the following: racism, migrants, islamophobia, antisemitism, religion, homophobia, sexism, ideology, media, politics, individual and other~\cite{novak2021imsypp}. 
Because this data set is focused on targeted identity groups, it is particularly suitable for this study. 
However, what is missing from this dataset for our objective is the identification of the social groups for each of the target categories in these hate speech examples.

To address this limitation, we manually curated a list of Dutch Social Group Terms (SGT) following the approach used by Davani et al.~\cite{davani2021improving} for English. %We perform this curation in the social context of  the Dutch social context into consideration. 
A naive approach of simply translating the English language SGT into Dutch would not suffice. For example, a social group `Moroccan', which is a prevalent minority group in the Netherlands, would not be a relevant social group in the United States. At the same time, some other social groups such as `Sikhs' are irrelevant for the hate speech context of the Dutch society and thus are excluded.
Additionally, due to complexities of Dutch grammar, we add variants of the SGTs to the list because nouns tend to have an adjective form in Dutch and vice versa. For example, the adjective `Nederlands' means `Dutch', while the word `Nederlander' is a noun which means a `Dutch person'. Also, since the Dutch language conjugates adjectives, we added one conjugation form to the SGT list, for instance, only `Marokkaans' and not `Marokkaanse', to keep it simple. We did not consider plural nouns for this reason. Following this approach, we curated a list of 85 SGTs, which we refer to as \textit{the Dutch SGT List}. The full list is available in Appendix~\ref{ap_categories_sgt}.
%\noindent 
%\subsection{Dataset Preprocessing}
After identifying the SGTs, we filtered the dataset and found that 2,649 posts contained at least one SGT from the Dutch SGT list, referred to as \emph{baseline data}, which we use for the next steps.
%denoted as $Train Data$. We used these comments for further experiments. 
In addition, we perform standard preprocessing tasks, including removing emojis and deleting extra spaces, special signs, commas, and full stops. 

%------------------------------------------------
\subsection{LLM-based Counterfactual Generation} 
%------------------------------------------------
Inspired by recent work~\cite{sen2023people,mishra2024llm}, we introduce two distinct prompting approaches for generating counterfactual sentences using large language models (LLMs); both approaches take posts from the baseline dataset as input.
The first approach, denoted as \emph{LLMdef}, operates implicitly by instructing the LLM to modify social group terms in a given input post. These modifications are based on various identity attributes, such as gender, race, class, sexuality, political affiliation, religion, education level, age, and profession, among others. In this approach, the model is expected to identify the relevant social group term within the input post and replace it with another term from the same category. The selection of the replacement term is left to the discretion of the LLM, allowing for a dynamic and context-aware generation of counterfactuals. This method generated 15,175 counterfactual posts.

The second approach, denoted as \emph{LLMlist}, builds upon the first method but introduces an explicit mechanism for social group term substitution. Instead of relying solely on the LLM's internal decision-making, we provide a predefined list of social group terms, which we have carefully curated and documented in Appendix~\ref{ap_categories_sgt}. The model is then directed to generate counterfactual posts by substituting the social group term in the input sentence with an alternative from this predefined list. This explicit specification ensures greater control over the counterfactual generation process and enables more systematic and interpretable modifications.
By employing these two approaches, we aim to explore the capabilities of LLMs in generating counterfactual statements that reflect variations in social identity attributes, facilitating a deeper understanding of biases, fairness, and representation in language models. This method generated 21,562 counterfactual posts.
\begin{table*}[!h]
    \centering
    \caption{Dataset Statistics (cnt: count; len: average length of sentence; ent: entropy of social group terms).} %\ainote{missing from the table: Dutch templates?
    \small
    \setlength{\tabcolsep}{4pt}
    %\begin{tabular}{c@{}|c@{}|c@{}|c@{}|c@{}|c@{}|c@{}|c@{}|c@{}|c@{}|c@{}|c@{}|c@{}|c@{}|c@{}|c@{}}
    \begin{tabular}{c|c|c|c|c|c|c|c|c|c|c|c|c|c|c|c}
    \hline
    & \multicolumn{3}{c}{\textbf{Baseline}} & \multicolumn{3}{|c}{\textbf{LLMdef}} & \multicolumn{3}{|c}{\textbf{LLMlist}} & \multicolumn{3}{|c}{\textbf{SLL}} & \multicolumn{3}{|c}{\textbf{MGS}} \\
    \cline{2-16}
    \textbf{Category} & \textbf{cnt} & \textbf{len} & \textbf{ent} & \textbf{cnt} & \textbf{len} & \textbf{ent} & \textbf{cnt} & \textbf{len} & \textbf{ent} & \textbf{cnt} & \textbf{len} & \textbf{ent} & \textbf{cnt} & \textbf{len} & \textbf{ent} \\
    \hline
    Appropriate  & 1011 & 34.9 & 4.4 & 5801 & 29.9 & 4.8  & 8273 & 28.8 & 6.8 & 15624 & 31.2 & 5.8 & 6580 & 26.2 & 4.6 \\
    Inappropriate & 260 & 37.1 & 4.8 & 1527 & 32.2 & 5.2 & 2211 & 29.7 & 7.2 & 6850 & 33.7 & 5.9 & 2863 & 27.1 & 3.9 \\
    Offensive & 1327 & 39.8 & 4.9 & 7549 & 36.4 & 5.3 & 10713 & 34.2 & 7.2 & 25663 & 38.6 & 5.9 & 10481 & 39.3 & 3.6 \\
    Violent & 46 & 37.3 & 3.5 & 298 & 34.5 & 4.0 & 365 & 36.1 & 5.4 & 967 & 29.8 & 5.8 & 469 & 30.5 & 3.6 \\
    \hline
    \end{tabular}
    \label{tab:data_stats}
\end{table*}

%------------------------------------------------
\subsection{Sentence Log-Likelihood (SLL)} 
%------------------------------------------------
Following previous work~\cite{nadeem2021stereoset,davani2021improving}, we generate counterfactuals considering the log-likelihood of the sentence, denoted as $SLL$. Equation (\ref{eq_log}) quantifies the log-likelihood ($\lg$(P(x))) of a sentence, where $x_0$, $x_1$, .... , $x_{i-1}$, $x_i$ refer to the words in a sentence $x$.
\begin{equation}
f(x) = \log(P(x)) = \sum_{i=1}^n \log P(x_i | x_0, x_1, .... , x_{i-1} )
\label{eq_log}
\end{equation}
 
Consider, for example, the sentence, ``all Moroccans should go back to their country.'' The SLL method assumes that the word `Moroccans' is more likely to be replaced by `Turks' than `Germans'. Turks is more likely to occur linguistically because, unlike Germans, it is a minority group that is often discriminated against in this way in Dutch society. 
%This method is thus intended to preserve the likelihood of a sentence by generating counterfactuals with an equal or higher log-likelihood than the original sentence~\cite{davani2021improving}.
For each of 2,649 comments, we substitute the SGT in the comment with each of the other SGTs in the Dutch SGT list to obtain potential counterfactual comments. We employ the pre-trained GPT-2 model to compute log-likelihood for each of these potential counterfactual substitutions. We consider only those counterfactual comments that have a higher or equal log-likelihood than the original comment. This method generated 49,104 counterfactual comments.

%\noindent 
%------------------------------------------------
\subsection{Manual Group Substitution (MGS)}
%------------------------------------------------
In line with previous works~\cite{yang2020generating,madaan2021generate}, we perturb SGTs based on the specific identity group and grammatical function in the sentences, specifically for Dutch. This method works with dictionaries, substituting an SGT with other SGTs that are present in the dictionary. The substitution process is mathematically defined as:
\begin{equation}
x_{cf} = \{ x' |\ x' \in substitute(x,D)\}    
\end{equation}
In this equation, $x_{cf}$ refers to the set with correctly generated counterfactuals. The equation uses $substitute(x,D)$ to replace the original SGT in a sentence with a counterfactual SGT based on its location in the dictionary. We curate this dictionary that contains several lists of SGTs based on whether the SGT is an adjective or noun and whether it belongs to one of the following identity groups: Nationality, skin colour, migrants, gender, sexuality, religion, age and ideology. For example, the word `woman' would belong to the gender/noun category in the defined dictionary and would, therefore, be replaced only with other gender/noun terms, `transgender' and `man' in this case. In this manner, the MGS method creates grammatically correct and likely counterfactual sentences. 
%This category substitution is based on identity group and grammar. The identity groups are sexuality, gender, nationality, migrants, religion, skin colour, ideology and age. These groups are further divided into either nouns or adjectives. This results in a total of 13 categories. For example, the category “gender\_terms\_noun” is used to substitute nouns that fall into the gender category. 
Following this process, we generate 20,393 counterfactual comments.

Table~\ref{tab:data_stats} describes these datasets in terms of count, average sentence length, and entropy of SGTs. % for original and all counterfactual datasets. \ainote{shoult there be 6 datasets, though? only 5 appear in the table. Also, would it make more sense to move this table earlier in the section? }

\begin{table}[!h]
    \centering
    \small
    \caption{Performance Metrics for Evaluated Models (Accuracy, F1 Score, Precision and Recall)}
    \begin{tabular}{c|c|c|c|c}
    \hline
    \textbf{Model} & \textbf{Acc} & \textbf{Prec} & \textbf{Rec} & \textbf{F1} \\
    \hline
    Baseline & 0.75 & 0.78 & 0.75 & 0.75 \\
    BERTje + LLMdef & 0.79 & 0.61 & 0.62 & 0.61 \\
    BERTje + LLMlist & 0.77 & 0.65 & 0.52 & 0.61 \\
    BERTje + SLL & 0.79 & 0.79 & 0.79 & 0.79 \\
    BERTje + MGS & 0.79 & 0.79 & 0.79 & 0.79 \\
    \hline
    \end{tabular}
    \label{tab:perf_metric}
\end{table}
\begin{table*}[!h]
    \centering
    \caption{Performance results of classification models. All models are the pretrained BERTje Dutch language model finetuned with the counterfactual datasets generated as described in Section~\ref{sec:counterf-data}. The baseline is the original BERTje model from~\cite{de2019bertje}.}
    \small
    \setlength{\tabcolsep}{4pt}
    \begin{tabular}{c|c|c|c|c|c|c|c|c|c|c|c|c|c|c|c}
    \hline
    & \multicolumn{3}{c}{\textbf{Baseline}} & \multicolumn{3}{|c}{\textbf{LLMdef}} & \multicolumn{3}{|c}{\textbf{LLMlist}} & \multicolumn{3}{|c}{\textbf{SSL}} & \multicolumn{3}{|c}{\textbf{MGS}} \\
    \cline{2-16}
    \textbf{Category} & \textbf{Prec} & \textbf{Rec} & \textbf{F1} & \textbf{Prec} & \textbf{Rec} & \textbf{F1} & \textbf{Prec} & \textbf{Rec} & \textbf{F1} & \textbf{Prec} & \textbf{Rec} & \textbf{F1} & \textbf{Prec} & \textbf{Rec} & \textbf{F1} \\
    \hline
    Appropriate  & 0.91 & 0.72 & 0.81 & 0.87 & 0.83 & 0.85 & 0.87 & 0.81 & 0.84 & 0.86 & 0.83 & 0.85 & 0.86 & 0.84 & 0.85 \\
    Inappropriate & 0.54 & 0.28 & 0.37 & 0.38 & 0.38 & 0.38 & 0.32 & 0.42 & 0.36 & 0.33 & 0.36 & 0.34 & 0.33 & 0.36 & 0.35 \\
    Offensive & 0.63 & 0.91 & 0.75 & 0.76 & 0.81 & 0.78 & 0.76 & 0.80 & 0.78 & 0.77 & 0.81 & 0.79 & 0.79 & 0.80 & 0.79  \\
    Violent & 0.66 & 0.44 & 0.53 & 0.42 & 0.44 & 0.43 & 0.43 & 0.47 & 0.45 & 0.58 & 0.44 & 0.50 & 0.55 & 0.49 & 0.52 \\
    \hline
    \end{tabular}
    \label{tab:per_class_results}
\end{table*}

%%%%%%%%%%%%%%%%%%%%%%%%%%%%%%%%%%%%%%%%%%%%%%
\section{Qualitative Assessment}
%%%%%%%%%%%%%%%%%%%%%%%%%%%%%%%%%%%%%%%%%%%%%%
%Besides quantitative measures, we also qualitatively evaluate the four counterfactual models. 
Next, we perform qualitative evaluation to assess whether \emph{realistic} counterfactuals are created. 
%Evaluating the counterfactually generated sentences sheds light on how realistic they are and, thus, the performance of these methods.\\
\subsection{Sentence Log-likelihood Evaluation}
%Analyzing counterfactual sentences that are generated by the SLL method reveals insights into the performance of this generation method. 
SLL methods do not create realistic counterfactuals. Firstly, some counterfactuals do not adhere to the Dutch grammar rules. Table~\ref{tab:cf_sll} in Appendix~\ref{ap_qualitative} shows an incorrectly generated counterfactual that uses ``jong'' as an adjective when Dutch grammar rules dictate that it should be ``jonge'' (because it belongs to a plural noun). This problem does not occur in English because adjectives are not conjugated based on the matching noun in this language. Therefore, this problem is specific to the Dutch context. In addition, Table~\ref{tab:cf_sll} demonstrates the problem of substituting the adjective ``Turkish'' when a noun is expected, according to Dutch grammar rules. Sentences that do not adhere to Dutch grammar are expected to get lower sentence likelihood since faulty grammar is expected to occur less often than correct grammar. However, this is not the case for the SLL method, suggesting the suboptimality of this method in the Dutch language regarding grammar adherence.
Secondly, the SLL method does not always generate counterfactuals that adhere to the sentence context. Table \ref{tab:cf_sll} shows the original sentence ``coming out as female''. A correctly generated example substitutes ``female'' with ``gay'', as this is a usual expression, retaining the sentence's meaning. However, the sentence likelihood method also substitutes the word ``female'' with ``young'', which is not a usual expression and, therefore, unrealistic. The counterfactual with ``young'' would have been expected to receive a lower log-likelihood than the original sentence, but this is not the case. This suggests that the method fails to properly process sentence context in Dutch.

\subsection{Manual Group Substitution Evaluation}
As compared to SLL, this method tends to adhere better to the Dutch grammar rules than the SLL method but still makes occasional mistakes. This happens especially when a word can be used as both an adjective and a noun, which is the case for several Dutch SGTs. Table~\ref{tab:cf_mgs} in Appendix~\ref{ap_qualitative} shows a sentence in which this is the case. The original sentence contains the word ``Chinese''. This word can be both an adjective and a noun in Dutch, resulting in the generation of grammatically incorrect sentences. In this case, when a noun is treated as an adjective and vice-versa, the method produces grammatically incorrect counterfactuals. The method is thus sub-optimal in creating realistic counterfactuals.
Moreover, the question remains whether MGS captures the context of a sentence adequately. Even when sentences adhere to Dutch grammar rules, the original meaning of the sentence is not fully captured by every counterfactual generated. Table~\ref{tab:cf_mgs} shows an example. The original sentence ``speak Dutch'' requires substituting words from the adjective list. The sentence correctly substitutes an adjective, but ``Latina'' is not a language, creating an unlikely sentence. %Therefore, MGS does not always create realistic counterfactuals.

\subsection{LLM Evaluation}
Generally, we see that both LLM methods can solve issues that arise in the other generation methods. MGS or SLL generation methods can create sentences like ``a black (een zwart)'', that are incorrect in the Dutch language. An LLM can work around this issue, by substituting ``a black person'', making the counterfactual sentences grammatically correct.
However, LLMlist (Table~\ref{tab:cf_llmlist} in Appendix~\ref{ap_qualitative}) still makes grammatical mistakes, usually based on already existing mistakes in the training data. The method creates sentences like ``marokkaan broeders'' rather than ``marokkaanse broeders'' from the original sentence ``mocro broeders'', which is grammatically incorrect. Additionally, this method creates unlikely counterfactuals and makes interpretation mistakes. They change, for example, ``iemand zwart maken'' (expression in Dutch, which means to discredit someone) to ``iemand wit maken'' which is incoherent. ``Black'' here is incorrectly detected as an SGT in this context. It seems that the LLMlist method strictly adheres to the specified group terms in the prompt when detecting SGTs and creating counterfactuals. This is further demonstrated by the creation of counterfactuals with a wide range of SGTs, despite being unlikely. For example, it will create a sentence like ``coming out as old'' rather than coming out as a specific gender or sexuality. These examples demonstrate the limits of using a pre-defined list within LLMs in the realm of counterfactual generation.
In contrast, LLMdef (Table~\ref{tab:cf_llmdef} in Appendix~\ref{ap_qualitative}) creates more realistic counterfactuals that fall within a specific social group. It will, for example, retain the gender aspect in a sentence like ``coming out as a woman'', creating only counterfactuals with a gender group term. This results in more likely counterfactuals. The qualitative analysis also shows a wider variety of social group terms than present in the SGT list, but can go beyond that, for example, by changing stereotypically Dutch names into Arabic-sounding or English sounding names. This strength can, however, work counterproductively. Too freely interchanging terms, creates counterfactuals that are too dissimilar from the original sentence. For example, LLMdef substituted the colors of a Dutch football club with ``education'' group terms, creating an unlikely counterfactual. Such \textit{free} substitution can result in either a mismatch between the generated sentence and the original label or an unlikely synthetic sentence that does not resemble human speech.

%%%%%%%%%%%%%%%%%%%%%%%%%%%%%%%%%%%%%%%%%%%%%%
\section{Performance Evaluation}% on Augmented Datasets}
%%%%%%%%%%%%%%%%%%%%%%%%%%%%%%%%%%%%%%%%%%%%%%
We are interested in understanding the counterfactual fairness of pre-trained models for Dutch hate speech detection and evaluating methods for improving it.
To this end, we focus on the BERTje model, a monolingual pre-trained model used in previous research in Dutch hate speech detection~\cite{markov2022role,novak2021imsypp}. 
This model categorizes hate speech into four classes: `acceptable', `inappropriate', `offensive', and `violent', corresponding to the categories in the IMSyPP dataset. The BERTje model was fine-tuned specifically for the IMSyPP dataset~\cite{de2019bertje}, making it a natural choice for evaluating fairness in this context.
Additionally, we further fine-tune the baseline BERTje model with counterfactual sentences generated using SLL, MGS and LLM-based methods. This results in five models: (1)~Baseline BERTje, (2)~BERTje + SLL, (3)~BERTje + MGS, (4)~BERTje + LLMdef, and (5)~BERTje + LLMlist, which will be evaluated on (counterfactual) fairness and performance. 
We tested all of these models on the test data from the IMSyPP dataset with stratified sampling. 

As shown in Table~\ref{tab:perf_metric}, the best-performing models are BERTje+SLL and BERTje+MGS, both achieving 79\% accuracy with better precision, recall, and F1-scores.
LLMdef and LLMlist do not give better precision and recall, suggesting that counterfactual data generated by LLMs might introduce noise.

Analyzing the performance per class label (Table \ref{tab:per_class_results}) gives more information on the model performance for specific hate speech classes. Interestingly, performance scores differ greatly between classes. The ``inappropriate'' and ``violent'' classes generally have lower precision, recall, and F1 scores than the ``appropriate'' and ``offensive'' classes, indicating class imbalances in the training data.
%as pointed out earlier in Table \ref{tab:dataset_stats}. 
The fine-tuned models show a slightly greater discrepancy than the baseline model.

%%%%%%%%%%%%%%%%%%%%%%%%%%%%%%%%%%%%%%%%%%%%%%
\section{Fairness Assessment}
%%%%%%%%%%%%%%%%%%%%%%%%%%%%%%%%%%%%%%%%%%%%%%
We also evaluate the fairness of these models using individual (counterfactual) and group fairness metrics. 
For fairness evaluation, we have categorized the SGTs into seven groups based on nationality, skincolor, gender, sexuality, religion, age, and ideology (see Table~\ref{tab:sgt_categories} in Appendix~\ref{ap_categories_sgt}).
For an evaluation that focuses on fairness, we create a set of similar counterfactual data, which we refer to as \textit{Dutch counterfactual evaluation dataset}. 
We manually curated this dataset with synthetic non-stereotypical instances created from a given set of templates. These templates consist of short toxic and non-toxic phrases like ``hug [identity term]'' or ``destroy [identity term]'', in which SGTs are substituted. 
The idea is to emphasize adjectives without explicitly connecting them to the substituted SGTs, thereby assessing the model's fairness based on adjective interpretation.
All sentences with four tokens or fewer were selected from the dataset of Dixon et al.~\cite{dixon2018measuring} and manually translated to Dutch. 
In this process, template sentences with equal translations were only added once. For instance, the words `terrible' and `horrible' both translate to `verschrikkelijk' in Dutch.
This resulted in 17 toxic and 17 non-toxic synthetic templates. Consequently, the Dutch SGTs were substituted into these sentence templates, resulting in this Dutch counterfactual dataset with 2,890 sentences.

Note that the models are trained on four labels (appropriate, inappropriate, offensive or violent), and are being evaluated on the Dutch counterfactual dataset which has toxic and non-toxic labels. 
Because of the discrepancy between the predicted labels and the original labels, we consider the `appropriate' and `inappropriate' labels as non-toxic and the `offensive' and `violent' as toxic.

\textbf{Counterfactual Fairness:} 
We use Counterfactual Token Fairness (CTF)~\cite{garg2019counterfactual} as a metric to assess the counterfactual fairness of our models. CTF aims to measure the fairness of the model outputs by assessing how much those outputs change when the inputs are altered with counterfactual examples. The CTF is quantified using the following equation:
\begin{equation}
    CTF(X,X_{cf}) = \sum_{x \in X} \sum_{x' \in X_{cf}} |g(x) - g(x')|
\end{equation}

In this equation, $X$ represents the set of original input instances, which have at least one of the SGTs. Each instance $x$ in $X$ has a corresponding counterfactual instance $x'$ in the set $X_{cf}$. 
The functions $g(x)$ and $g(x')$ compute the labels for the original sentence and the corresponding counterfactual sentence, respectively.
%logits for an input $x$, which are then passed through a classifier function $f = \sigma(g(x))$, where $\sigma$ represents a \textit{softmax} function. 
The absolute difference $|g(x) - g(x')|$ captures how much the model's predictions change between an original instance $x$ and its counterfactual $x'$. A lower CTF value indicates that the model's outputs for original sentences and their counterfactual versions are more similar. A lower CTF is desired because it suggests that the model treats counterfactual variants of the same input fairly.

Analyzing and comparing the CTF of the models clarifies how the specified counterfactual generation methods affect counterfactual fairness. As shown in Table~\ref{tab:cf_results}, the baseline model has 0.24 as an overall. All counterfactual models improve fairness for non-toxic class but their performance deteriorates for toxic class. 
%MGS reaches a score of 0.14 and SLL reaches a score of 0.10. 
Among the counterfactual models, SLL is the most counterfactually fair both on average and also for the toxic templates. The baseline model gives more biased predictions for non-toxic templates compared to toxic templates.

\begin{table}[htbp]
    \centering
    \small
    \caption{Counterfactual fairness results. Lower values mean
    more fair model.}
    \setlength{\tabcolsep}{4pt}
    \begin{tabular}{c|c|c|c}
    \hline
    \textbf{Model} & \textbf{Toxic} & \textbf{Non-Toxic} & \textbf{Average}\\
    \hline
      Baseline & 0.11 & 0.36 & 0.24 \\
      BERTje+LLMdef & 0.26  & 0.011 & 0.13 \\
      BERTje+LLMlist & 0.32  & 0.001 & 0.16 \\
      BERTje+SLL & 0.20 & 0.001 & 0.10 \\
      BERTje+MGS & 0.28 & 0.003 & 0.14 \\
    \hline
    \end{tabular}
    \label{tab:cf_results}
\end{table}
\begin{table*}[h]
    \centering
    \small
    \setlength{\tabcolsep}{4pt}
    \caption{CTF Scores per Social Category. Lower values mean more fair model.}
   \begin{tabular}{c|c|c|c|c|c|c|c|c|c|c|c}
    \hline
    \textbf{Model} & \multicolumn{2}{c|}{\textbf{Baseline}} & \multicolumn{2}{c|}{\textbf{LLMdef}} & \multicolumn{2}{c|}{\textbf{LLMlist}} & \multicolumn{2}{c|}{\textbf{SLL}} & \multicolumn{2}{c}{\textbf{MGS}} \\  
    \hline
                      & \textbf{Tox} & \textbf{NonTox}  & \textbf{Tox} & \textbf{NonTox} & \textbf{Tox} & \textbf{NonTox} & \textbf{Tox} & \textbf{NonTox} & \textbf{Tox} & \textbf{NonTox} \\
    \hline
    Religion          & 0.02  & 0.38  & 0.20  & 0.00    & 0.23 & 0.00 & 0.22 & 0.00 & 0.25 & 0.00 \\
    Skin Color        & 0.31  & 0.26  & 0.14  & 0.00    & 0.25 & 0.00 & 0.11 & 0.00 & 0.27 & 0.00 \\
    Nationality       & 0.10  & 0.37  & 0.25  & 0.02    & 0.33 & 0.00 & 0.20 & 0.00 & 0.24 & 0.00 \\
    Ideology          & 0.01  & 0.46  & 0.39  & 0.01    & 0.27 & 0.01 & 0.19 & 0.01 & 0.33 & 0.01 \\
    Age               & 0.07  & 0.03  & 0.08  & 0.00    & 0.17 & 0.00 & 0.10 & 0.00 & 0.10 & 0.00 \\
    Gender            & 0.10  & 0.07  & 0.21  & 0.00    & 0.23 & 0.00 & 0.22 & 0.00 & 0.25 & 0.00 \\
    Sexuality         & 0.05  & 0.43  & 0.17  & 0.00    & 0.23 & 0.00 & 0.18 & 0.00 & 0.29 & 0.02 \\
    \hline
    \end{tabular}
    \label{tab:ctf_combined}
\end{table*}

Table~\ref{tab:ctf_combined} presents CTF scores for each SGT category for different models. 
As evident, all counterfactual models perform well for non-toxic templates for each of the SGT categories. 
For the toxic templates, counterfactual models match the fairness with base model only for the age category. 
In all remaining categories, fairness performance of counterfactual models underperform when compared to the baseline model.

\begin{table}[!h]
    \centering
    \small
    \setlength{\tabcolsep}{4pt}
    \caption{Fairness Metrics for fine-tuned models (based on groups rather than group terms). We compute demographic parity difference (DPD) and equalized odds difference (EOD). Lower values mean more fair model.}
    \begin{tabular}{c|c|c}
       \hline
       \textbf{Model}  & \textbf{DPD} & \textbf{EOD} \\
       \hline
       Baseline & 0.38 & 0.53 \\
       BERTje+LLMdef & 0.09  & 0.18 \\
       BERTje+LLMlist & 0.13 & 0.25 \\
       BERTje+SLL & 0.06 & 0.11 \\
       BERTje+MGS & 0.18 & 0.36 \\
       \hline 
    \end{tabular}
    \label{tab:group_fairness}
\end{table}
%\subsection
\textbf{Group Fairness:} Comparing the group fairness~\cite{hardt2016equality} results gives insight into how the specified counterfactual generation methods impact a model’s group fairness~(Table \ref{tab:group_fairness}). The baseline model has the highest demographic parity difference (DPD) of 0.38, indicating that the model’s predicted positive rate differs greatly among identity groups. All counterfactual models perform better than the baseline. MGS improves the score to 0.18, while the SLL method reaches the lowest DPD of 0.06. Both MGS and SLL perform better than LLM-generated counterfactuals.  
Furthermore, the baseline model scores poorly for the equalized odds difference (EOD) with 0.53 as maximum score. 
The LLMdef and LLMlist methods improve this score to 0.18 and 0.25, respectively. MGS method moves it to 0.36, while SLL performs the best with  the lowest EOD score of 0.11, indicating same trend as DPD. 
Although these scores are still considered sub-optimal, they signify an improvement with respect to the baseline model. 
Overall, both counterfactual generation models significantly improve in DPD and EOD, indicating fairer, less biased models. The SLL method outperforms MGS for both DPD and EOD, implying superior group fairness.

%%%%%%%%%%%%%%%%%%%%%%%%%%%%%%%%%%%%%%%%%%%%%%
\section{Conclusions and Future Work}
%%%%%%%%%%%%%%%%%%%%%%%%%%%%%%%%%%%%%%%%%%%%%%
This paper generates and evaluates the effect of counterfactual Dutch-language datasets for hate speech detection. 
The research finds that counterfactual generation methods introduce notable fairness improvements while maintaining performance in the Dutch context.
The resulting datasets are publicly available for research~\footnote{https://github.com/Juulba/Dutch-counterfactual-fairness}, being the first such datasets for the Dutch language.
%Four of these datasets (two generated using previously proposed techniques applied to English language datasets, and two that use LLMs) are evaluated in terms of performance improvements to classification of hate speech instances and counterfactual token fairness. 
Our empirical evaluations indicate improved overall counterfactual fairness and classification performance. The counterfactual models, in general, slightly improve on F1 score, precision and recall. Performance improvements may be partially due to increased training data for the counterfactual models, yet sometimes increased training data also exacerbates class imbalances. However, performing ablation studies with varying size of training data and fine-tuning strategies can be done in the future to study the impact of data variations on performance and fairness. In addition, extending this work with state-of-the-art LLMs can provide different results and insights.

We find that some counterfactuals generated using the SLL and MGS methods do not adhere to the Dutch grammar rules, and sometimes the generated sentences do not make sense in Dutch. In contrast, LLM-generated counterfactuals tend to be closer to correct Dutch. 
However, the Dutch-language BERTje model finetuned on SLL and MGS-based counterfactual datasets leads to improved fairness, while finetuning on the LLM-generated datasets does not lead to better fairness, particularly for the toxic class.
Interestingly, counterfactual models also positively impact group fairness metrics (DPD and EOD).

However, results show that counterfactual token fairness is worse (i.e., it increases) for the toxic class. This result is counterintuitive; it indicates that the counterfactual generation methods might introduce new biases or enforce already-existing ones that disproportionately affect the CTF of the toxic class. These biases might be introduced by generating unrealistic counterfactuals, which introduce noise into the data set. However, more research is needed to understand this phenomenon.

\bibliography{main}

\newpage

\appendix

%\section{Example Appendix}
%\label{sec:appendix}

%\appendix

%\section{Social Group Terms, Dutch Context}
%%%
%\label{app_sgt}
%SGT List
%Heteroseksueel, hetero, homoseksueel, homo, trans, transgender, non-binair, vrouw, vrouwelijk, man, mannelijk, masculien, biseksueel, lesbisch, bi, intersekse, turk, turks, marokkaan, marokkaans, mocro, nederlander, nederlands, hollander, hollands, indonesisch, indo, suri, surinamer, surinaams, antiliaan, antilliaans, zwart, wit, blank, syrisch, syrier, chinees, duits, duitser, belg, belgisch, afrikaans, afrikaan, jood, joods, asielzoeker, vluchteling, autochtoon, allochtoon, engels, aziatisch, aziaat, amerikaan, amerikaans, arabier, arabisch, latina, latino, immigrant, migrant, donker, christelijk, katholiek, moslim, protestant, rechts, links, marxist, marxistisch, populist, populistisch, feminist, feministisch, liberaal, boer, communist, communistisch, oudere, oud, jongere, jong, volwassene, volwassen, student.

\section{Categories of Social Group Terms in the Dutch Context}
\label{ap_categories_sgt}
Table~\ref{tab:sgt_categories} maps the social group terms from the Dutch context into seven categories. 
\begin{table}[!ht]
\centering
\caption{Categorization of Social Group Terms.}
\begin{tabular}{|c|p{5cm}|}
\hline
\textbf{Category} & \textbf{Values} \\
\hline
nationality & turk, marokkaan, mocro, nederlander, hollander, indo, suri, surinamer, antiliaan, syrier, chinees, duitser, belg, afrikaan, aziaat, amerikaan, arabier, latina, latino, turks, marokkaans, mocro, nederlands, hollands, indonesisch, indo, suri, surinaams, antilliaans, syrisch, chinees, duits, belgisch, afrikaans, engels, aziatisch, amerikaanse, arabisch, latina, latino, immigrant, migrant, asielzoeker, vluchteling, autochtoon, allochtoon \\
\hline
skincolor & wit, blank, zwart, donker \\
%\hline
%migrant & asielzoeker, vluchteling, autochtoon, allochtoon, immigrant, migrant \\
\hline
gender & vrouw, man, transgender, non-binair, vrouwelijk, mannelijk, masculien, intersekse \\
\hline
sexuality & homo, hetero, heteroseksueel, homoseksueel, trans, bi, lesbisch, biseksueel \\
\hline
religion & christelijk, katholiek, moslim, protestant, jood, joods \\
\hline
age & oudere, jongere, volwassene, student, oud, jong, volwassen \\
\hline
ideology & marxist, populist, feminist, liberaal, boer, communist, rechts, links, marxistisch, populistisch, feministisch, liberaal, communistisch \\
\hline
\end{tabular}
\label{tab:sgt_categories}
\end{table}

%%%%%%%%%%%%%%%%%%%%%%%%%%%%%%%%%%%%%%
\section{LLM Prompts}
%%%%%%%%%%%%%%%%%%%%%%%%%%%%%%%%%%%%%%
We use the following two LLM prompts for generating counterfactual sentences.

%------------------------------------------------
%\subsection{LLMdef}
%------------------------------------------------
\textbf{LLMdef}: \emph{Your task is to generate five counterfactual sentences for a given input sentence in Dutch. Identify social group terms in the input sentence and modify the input sentence by changing the social group(s) with other social group terms. 
Each counterfactual sentence should differ from the original while preserving the original meaning. 
Social group terms are terms that signify social groups based on characteristics like gender, race, class, sexuality, political leaning, religion, education, age, profession etc. 
Please generate five distinct counterfactual sentences for the input sentence: $input\_sentence$
Your response should be a JSON list of five dictionaries where each dictionary contains: 
- `input sentence': the original sentence.
- `counterfactual sentence': the generated counterfactual sentence.  
}
%\subsection{LLMlist}

\textbf{LLMlist:} \emph{Your task is to generate seven counterfactual sentences for a given input sentence in Dutch language. 
Identify and modify the input sentence by changing the social group term(s) with other social group term(s) and change other words without changing the level of toxicity expressed in input sentence based on $input\_sentence$. 
The generated sentence should make contextual sense.
Some social group terms in Dutch context are: "heteroseksueel", "hetero", "homoseksueel", "homo", "trans", "transgender", "non-binair", "vrouw",
"vrouwelijk", "man", "mannelijk", "masculien", "biseksueel", "lesbisch", "bi",
"intersekse", "turk", "turks", "marokkaan", "marokkaans", "mocro", "nederlander", "nederlands", "hollander",
"hollands", "indonesisch", "indo", "suri", "surinamer", "surinaams", "antiliaan", "antilliaans", "zwart", "wit", "blank",
"syrisch", "syrier", "chinees","duits", "duitser", "belg", "belgisch", "afrikaans", "afrikaan", "jood",
"joods", "asielzoeker", "vluchteling", "autochtoon", "allochtoon", "engels", "aziatisch", "aziaat", "amerikaan",
"amerikaans", "arabier", "arabisch", "latina", "latino", "immigrant", "migrant", "donker", "christelijk",
"katholiek", "moslim", "protestant", "rechts", "links",
"marxist", "marxistisch", "populist", "populistisch", "feminist", "feministisch", "liberaal",
"boer", "communist", "communistisch", "oudere", "oud",
"jongere", "jong", "volwassene", "volwassen", and "student". Input sentence is: $input\_sentence$. 
Give your response as a dictionary of "counterfactual sentences" with consistent formatting.
}

\section{Dataset Statistics}
In Tables~\ref{tab:train_data},~\ref{tab:llm_def_data},~\ref{tab:llm_list_data},~\ref{tab:sll_data}, and ~\ref{tab:mgs_data}, we provide distribution of categories of social group terms in each of the four labels in all datasets.
\begin{table*}[!h]
\centering
\small
\caption{Baseline Dataset.}
\begin{tabular}{r|r|rrrrrrr}
\hline
label & count & age & gender & ideology &  nationality & religion & sexuality & skincolor\\
\hline
Appropriate & 1057 & 98 & 507 & 120 & 134 & 46 & 25 & 118 \\
Inappropriate & 289 & 17 & 72 & 16 & 83 & 36 & 20 & 41 \\
Offensive & 1436 & 66 & 503 & 259 & 298 & 86 & 64 & 156 \\
Violent & 47 & 3 & 19 & 5 & 11 & 2 & 3 & 4 \\
\hline
& 2825 & 184 & 1101 & 400 & 526 & 170 & 112 & 319 \\
\hline
\end{tabular}
\label{tab:train_data}
\end{table*}

\begin{table*}[!h]
\centering
\small
\caption{LLMdef generated Dataset.}
\begin{tabular}{r|r|rrrrrrr}
\hline
label & count & age & gender & ideology & nationality & religion & sexuality & skincolor \\
\hline
Appropriate & 6110 & 500 & 2574 & 633 & 1359 & 265 & 105 & 619 \\
Inappropriate & 1702 & 97 & 340 & 73 &  691 & 185 & 91 & 200 \\
Offensive & 8249 & 383 & 2464 & 1458 & 2388 & 459 & 262 & 800 \\
Violent & 304 & 16 & 98 & 46 & 99 & 10 & 15 & 20 \\
\hline
& 16305 & 996 & 5476 & 2210 & 4537 & 919 & 473 & 1639 \\
\hline
\end{tabular}

\label{tab:llm_def_data}
\end{table*}

\begin{table*}[!h]
\centering
\small
\caption{LLMlist generated Dataset.}
\begin{tabular}{r|r|rrrrrrr}
\hline
label & count & age & gender & ideology &  nationality & religion & sexuality & skincolor \\
\hline
Appropriate & 7965 & 1066 & 1971 & 636 & 2088 & 524 & 1022 & 580 \\
Inappropriate & 2277 & 189 & 429 & 144 & 793 & 168 & 323 & 189 \\
Offensive & 10747 & 965 & 2346 & 1483 & 3107 & 678 & 1251 & 787 \\
Violent & 360 & 29 & 109 & 26 & 98 & 24 & 55 & 17 \\
\hline
& 21160 & 2249 & 4855 & 2289 & 6086 & 1394 & 2651 & 1573 \\
\hline
\end{tabular}
\label{tab:llm_list_data}
\end{table*}

\begin{table*}[!h]
\centering
\small
\caption{SLL generated Dataset.}
\begin{tabular}{r|r|rrrrrrr}
\hline
label & count & age & gender & ideology &  nationality & religion & sexuality & skincolor\\
\hline
Appropriate & 15624 & 1652 & 1572 & 2586 & 5547 & 543 & 1956 & 1514 \\
Inappropriate & 6850 & 723 & 676 & 1111 & 2653 & 277 & 736 & 555 \\
Offensive & 25663 & 2731 & 2690 & 4110 & 9509 & 948 & 2922 & 2322 \\
Violent & 967 & 106 & 102 & 156 & 342 & 33 & 116 & 92 \\
\hline
& 49104 & 5212 & 5040 & 7963 & 18051 & 1801 & 5730 & 4483 \\
\hline
\end{tabular}
\label{tab:sll_data}
\end{table*}

\begin{table*}[!h]
\centering
\small
\caption{MGS generated Dataset.}
\begin{tabular}{r|r|rrrrrrr}
\hline
label & count & age & gender & ideology &  nationality & religion & sexuality & skincolor \\
\hline
Appropriate & 6580 & 198 & 894 & 518 & 4532 & 117 & 133 & 188 \\
Inappropriate & 2863 & 28 & 110 & 48 & 2418 & 96 & 105 & 58 \\
Offensive & 10481 & 133 & 818 & 1302 & 7506 & 213 & 287 & 222 \\
Violent & 469 & 4 & 36 & 33 & 361 & 6 & 21 & 8 \\
\hline
 & 20393 &	363	& 1858 &	1901	& 14817 &	432	& 546	& 476 \\
\hline
\end{tabular}
\label{tab:mgs_data}
\end{table*}

\section{Qualitative Assessment}
\label{ap_qualitative}
\begin{table*}
    \centering
    \caption{Examples of incorrect counterfactual sentences generated by SLL method. English translation is within brackets.}
    \begin{tabular}{p{4.5cm}|p{7.5cm}}
    \hline
       \multicolumn{2}{c}{\textit{incorrect adjective conjugation}} \\
       \hline
       \textbf{Original sentence} & volwassen mensen (adult people) \\
       \textbf{Incorrect counterfactual} & jong mensen (young people) \\
       \textbf{Correct counterfactual} & jongere mensen (younger people) \\
    \hline
    \multicolumn{2}{c}{\textit{wrong noun adjective substitution}} \\
       \hline
       \textbf{Original sentence} & zielige allochtoon (pitiful immigrant) \\
       \textbf{Incorrect counterfactual} & zielige Turks (pitiful Turkish) \\
       \textbf{Correct counterfactual} & zielige Turk (pitiful Turk) \\
    \hline
    \multicolumn{2}{c}{\textit{unlikely counterfactual generation}} \\
       \hline
       \textbf{Original sentence} & uit de kast komen als vrouw (coming out as female) \\
       \textbf{Incorrect counterfactual} & uit de kast komen als jong (coming out as young) \\
       \textbf{Correct counterfactual} & uit de kast komen als homo (coming out as gay) \\
    \hline
    \end{tabular}
    \label{tab:cf_sll}
\end{table*}

\begin{table*}
    \centering
    \caption{Examples of incorrect counterfactual sentences generated by MGS method. English translation is within brackets.}
    \begin{tabular}{p{4.5cm}|p{7.5cm}}
    \hline
       \multicolumn{2}{c}{\textit{wrong noun adjective substitution}} \\
       \hline
       \textbf{Original sentence} & Chinees gehaald (taken out Chinese) \\
       \textbf{Incorrect counterfactual} & Turk gehaald (taken out Turk) \\
       \textbf{Correct counterfactual} & Turks gehaald (taken out Turkish) \\
    \hline
    \multicolumn{2}{c}{\textit{unlikely counterfactual generation}} \\
       \hline
       \textbf{Original sentence} & spreek Nederlands (speak Dutch) \\
       \textbf{Incorrect counterfactual} & spreek Latina (speak Latina) \\
       \textbf{Correct counterfactual} & spreek Belgisch (speak Belgian) \\
    \hline
    \end{tabular}
    \label{tab:cf_mgs}
\end{table*}

\begin{table*}
    \centering
    \caption{Examples of incorrect counterfactual sentences generated by LLMlist. English translation is within brackets.}
    \begin{tabular}{p{4.5cm}|p{7.5cm}}
    \hline
       \multicolumn{2}{c}{\textit{incorrect adjective conjugation LLMlist}} \\
       \hline
       \textbf{Original sentence} & Mocro broeders (Moroccon brothers) \\
       \textbf{Incorrect counterfactual} & Marokkaan broeders (Moroccan person brothers) \\
       \textbf{Correct counterfactual} & Marokkaanse broeders (Moroccon brothers) \\
    \hline
    \multicolumn{2}{c}{\textit{unlikely counterfactual generation}} \\
       \hline
       \textbf{Original sentence} & iemand zwart maken (discredit someone) \\
       \textbf{Incorrect counterfactual} & iemand wit maken (to make someone white) \\
    \hline
    \multicolumn{2}{c}{\textit{unlikely counterfactual generation}} \\
       \hline
       \textbf{Original sentence} & uit de kast komen als vrouw (coming out as female) \\
       \textbf{Incorrect counterfactual} & uit de kast komen als jong (coming out as old) \\
       \textbf{Correct counterfactual} & uit de kast komen als man (coming out as male) \\
    \hline
    \end{tabular}
    \label{tab:cf_llmlist}
\end{table*}

\begin{table*}
    \centering
    \caption{Example of incorrect counterfactual sentence generated by LLMdef. English translation is within brackets.}
    \begin{tabular}{p{4.5cm}|p{7.5cm}}
    \hline
    \multicolumn{2}{c}{\textit{unlikely counterfactual generation}} \\
       \hline
       \textbf{Original sentence} & rood, zwart en groen (red, black and green) \\
       \textbf{Incorrect counterfactual} & student, docent en directeur (student, teacher and principal) \\
    \hline
    \end{tabular}
    \label{tab:cf_llmdef}
\end{table*}

\end{document}